\documentclass[10pt,twocolumn,letterpaper]{article}
\usepackage[pagenumbers]{cvpr} 

\definecolor{cvprblue}{rgb}{0.21,0.49,0.74}
\definecolor{methodgroup}{RGB}{220,235,255}

\usepackage{xcolor}
\usepackage{ifthen}

\usepackage[pagebackref,breaklinks,colorlinks,allcolors=cvprblue]{hyperref}
\usepackage{multirow}
\usepackage{bbding} 
\usepackage{algorithm}
\usepackage{algpseudocode}
\usepackage{array}
\usepackage[table]{xcolor} 

\newcommand{\name}{Nav-$R^2$}
\newcommand{\mname}{SA-Mem}

\newcommand{\dname}{\name: Dual‑Relation Reasoning for Generalizable Open‑Vocabulary Object‑Goal Navigation}
\newcommand{\dmname}{Similarity-Aware Memory}

\title{\dname}

\author{Wentao Xiang$^{1}$$^{,}$$^{2}$, Haokang Zhang$^{1}$, Tianhang Yang$^{1}$, \\ 
Zedong Chu$^{2}$$^{\dagger}$, Ruihang Chu$^{1}$, Shichao Xie$^{2}$, 
Yujian Yuan$^{2}$, Jian Sun$^{2}$, Zhining Gu$^{2}$, \\ 
Junjie Wang$^{1}$, 
Xiaolong Wu$^{2}$, Mu Xu$^{2}$ and Yujiu Yang$^{1}$$^{\dagger}$  \\
\textbf{$^{1}$} Tsinghua University  \quad 
\textbf{$^{2}$} Amap, Alibaba Group. \\
{\tt\small thu\_xiangwentao@163.com,} 
{\tt\small chuzedong.czd@alibaba-inc.com,} 
{\tt\small yang.yujiu@sz.tsinghua.edu.cn}
}

\begin{document}
\maketitle

\let\thefootnote\relax\footnotetext{
\noindent \quad $^\dagger$ Corresponding author.}

\begin{figure*}[h]
    \centering
    \includegraphics[width=1\textwidth, 
        trim=0in 1in 0in 1in, clip
    ]{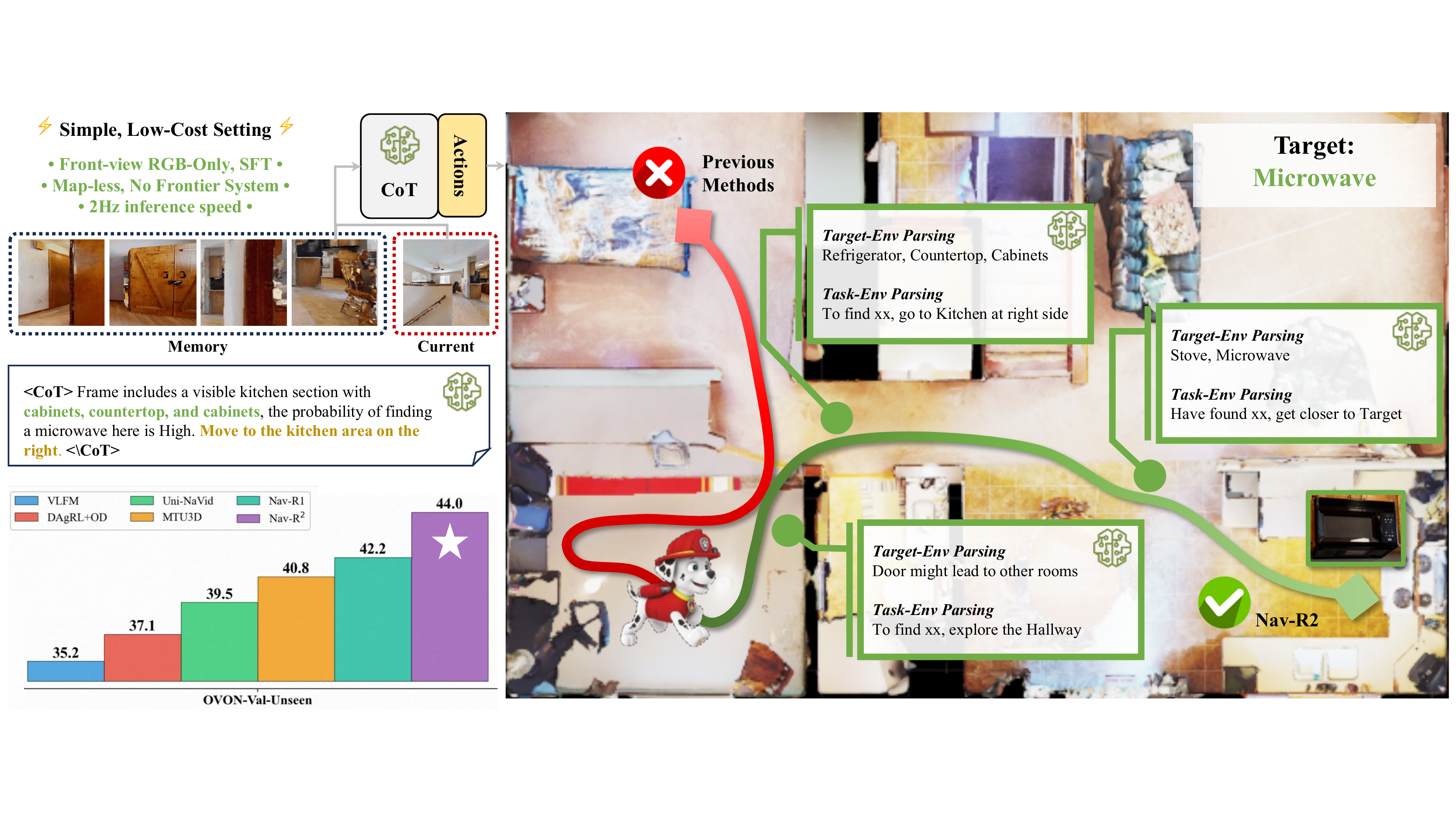}
    \caption{Overview of \name\ during an object-goal navigation episode (target: Microwave) in an unseen indoor environment and the results on OVON-val-unseen datasets. At each step, the agent parses the current observation to extract two core relations among three key entities: the environment–task relation (e.g., linking the navigation task to relevant regions such as kitchens or hallways) and the environment–target relation (e.g., associating visible objects such as cabinets or countertops with the likelihood of containing the target.}
    \label{fig:teaser}
\end{figure*}

\begin{abstract}

Object-goal navigation in open-vocabulary settings requires agents to locate novel objects in unseen environments, yet existing approaches suffer from opaque decision-making processes and low success rate on locating unseen objects.
To address these challenges, we propose \name, a framework that explicitly models two critical types of relationships, target-environment modeling and environment-action planning, through structured Chain-of-Thought (CoT) reasoning coupled with a \dmname.
We construct a Nav$R^2$-CoT dataset that teaches the model to perceive the environment, focus on target-related objects in the surrounding context and finally make future action plans.
Our \mname\  preserves the most target-relevant and current observation-relevant features from both temporal and semantic perspectives by compressing video frames and fusing historical observations, while introducing no additional parameters.
Compared to previous methods, \name\ achieves state-of-the-art performance in localizing unseen objects through a streamlined and efficient pipeline, avoiding overfitting to seen object categories while maintaining real-time inference at 2Hz. Resources will be made publicly available at
\href{https://github.com/AMAP-EAI/Nav-R2}{github link}.

\end{abstract}

\section{Introduction}
\label{sec:intro}

Object-goal navigation (ObjectNav) represents a fundamental challenge in embodied AI, requiring an agent to locate a specified target object in previously unseen indoor environments. In open-vocabulary settings, the agent must generalize to novel object categories and unfamiliar layouts without prior exposure, which demands robust perception, reasoning, and decision-making in diverse conditions. During navigation, robots must jointly reason over environmental perception and the given task to determine where to search, why a decision is made, and which direction to proceed. 

This process can be viewed as reasoning over two critical relationships: the \textbf{\textit{target-environment}} relation and the \textbf{\textit{environment-action}} relation. We define the target-environment relation as a perception-centric challenge, requiring the agent to reason about which objects and features in the observations are semantically or spatially relevant to the navigation target. In contrast, the environment-action relation is planning-centric, focusing on how the agent should translate its environmental understanding into future actions. We argue that explicitly and structurally modeling these two distinct reasoning axes is crucial for effective navigation, yet this remains a gap in existing research.

Some recent ObjectNav works~\citep{chaplot2020objectgoalnavigationusing, zhang2018semantic, majumdar2022zson, sethian1996fast, zhang2024uni@uninavid, zhang2025embodiednavigationfoundationmodel@navfom, lian2024tdanet} are incapable of such granular reasoning, implementing navigation through reinforcement learning or modular map construction. Other works~\citep{lin2025navcot@navcot, cai2025clcotnavclosedloophierarchicalchainofthought@clcotnav, liu2025nav@navr1, qiao2025opennavexploringzeroshotvisionandlanguage@opennav, cao2025cognav@cognav} emphasize interpretability via chain-of-thought (CoT) mechanisms. However, their reasoning structures are often implicit and structureless, failing to explicitly disentangle the perception-focused \textit{target-environment} reasoning from the planning-focused \textit{environment-action} reasoning. This limits their ability to robustly find target objects, especially unseen ones. Furthermore, while some works in video tasks~\citep{cheng2022xmemlongtermvideoobject, shi2025memoryvlaperceptualcognitivememoryvisionlanguageaction} focus on memory design, both compressing dense visual information to fewer tokens and abstracting it via text can cause information loss. This loss is particularly detrimental to modeling both relations, which rely on rich, real-time perceptual cues.

In contrast, we propose an object-goal navigation framework that explicitly models the target-environment and environment–action relationships within a unified CoT reasoning paradigm, while operating solely on first-person RGB input. To this end, we construct a dedicated CoT dataset tailored for open-vocabulary ObjectNav, enabling our \name\ to jointly reason about visual observations, task specifications, and historical spatial context through a \dmname(\mname). Our model, \name, is trained exclusively via supervised fine-tuning on simulated data, without requiring map construction, real-world data, or reinforcement learning. This design yields strong generalization to unseen object categories, supports inference at over 2Hz, and achieves state-of-the-art performance (success rate 44.0\%) on the open-vocabulary ObjectNav benchmark OVON~\citep{yokoyama2024hm3d@hm3dovon} val-unseen dataset.

In summary, our contributions are threefold:

(1) We propose a relational reasoning framework for object-goal navigation that explicitly models the \textit{target-environment} (perception) and \textit{environment–action} (planning) relationships, integrating this structured reasoning in a streamlined pipeline without introducing additional model parameters.

(2) We construct a novel Chain-of-Thought dataset specifically designed for training a generalizable object-goal navigation model capable of reasoning and modeling both two relationships.

(3) We develop \name, a vision-language reasoning model just trained via supervised fine-tuning on first-person RGB frames, achieving state-of-the-art performance in open-vocabulary ObjectNav and real-time inference at around 2Hz.
\section{Related Work}
\label{sec:formatting}

\subsection{Object Navigation}
Many progress has been made in object navigation tasks, and some of them are incapable of reasoning\citep{chaplot2020objectgoalnavigationusing, zhang2018semantic, majumdar2022zson, sethian1996fast, zhang2024uni@uninavid, zhang2025embodiednavigationfoundationmodel@navfom, lian2024tdanet, wu2024voronavvoronoibasedzeroshotobject@voronav, wortsman2019learning} implementing navigation through re-inforcement learning, modular map construction, while some other works emphasize interpretability via chain-of-thought mechanisms and structured reasoning frameworks~\citep{lin2025navcot@navcot, cai2025clcotnavclosedloophierarchicalchainofthought@clcotnav, liu2025nav@navr1, qiao2025opennavexploringzeroshotvisionandlanguage@opennav, cao2025cognav@cognav}. For example, Open-Nav \citep{qiao2025opennavexploringzeroshotvisionandlanguage@opennav} introduces a spatial-temporal CoT with open-source LLMs for continuous VLN scenarios; NavCoT \citep{lin2025navcot@navcot} proposes a trainable, disentangled CoT for VLN, separating future imagination, visual filtering, and action prediction; CL-CoTNav \citep{cai2025clcotnavclosedloophierarchicalchainofthought@clcotnav} applies hierarchical CoT to ObjectNav, reasoning over object-object and room-level relations with a closed-loop confidence mechanism; and VoroNav \citep{wu2024voronavvoronoibasedzeroshotobject@voronav} integrates spatial topology and scene descriptions to guide LLM waypoint selection.

\subsection{Memory Design}
Due to the nature of long sequence in video tasks and object navigation tasks, it is vital to store and update the historical observations. STM\citep{oh2019videoSTM}~constructs a memory bank for each object in the video, and matches every query frame to the memory bank to perform ``memory readout``, having been applied in neumerous works\citep{cheng2021mivos, seong2020kernelizedMemory, lu2020videoGraphMem, li2020fastGlobalContext, hu2021learning, Liang2020AFBURR, wang2021swiftnet, lai2020mast}. Memory are merged and updated to limit the maximum length and augment the feature quality through exponential moving averages, least-recently used strategy and so on. There also exist other methods\citep{cheng2022xmemlongtermvideoobject, shi2025memoryvlaperceptualcognitivememoryvisionlanguageaction} maintaining three independent yet deeply-connected feature memory stores: a rapidly updated \emph{sensory memory}, a high-resolution \emph{working memory}, and a compact thus sustained \emph{long-term memory} similar to the Atkinson–Shiffrin memory model~\citep{atkinson1968human}.

Despite notable progress, \textbf{critical limitations remain in Object-goal Navigation tasks}: (1)~the reasoning structures are implicit and unable to find target objects very well especially unseen objects, without explicit modeling of \textit{how target objects relate to environments and how environments relate to future action planning;} (2)~the memory lacks of interactions between various feature embeddings and interpretability. Consequently, inspired but different from these methods, our work introduces a ObjectNav CoT dataset and a framework consisting of \mname\ tailored for Open-Vocabulary Object Navigation Tasks, enabling explicit modeling of target-environment and environment-action relationships through RGB-only observations.

\section{Method}
This section introduces how the CoT dataset is constructed in \cref{subsec:dataset_construction_and_curation} and how our \name\ with \mname\ is built in \cref{subsec:\name}.

\subsection{Dataset Construction and Curation}
\label{subsec:dataset_construction_and_curation}




\begin{figure}[t]
    \centering
    \includegraphics[width=0.45\textwidth, 
        trim=3.8in 0in 3.7in 0in, clip
    ]{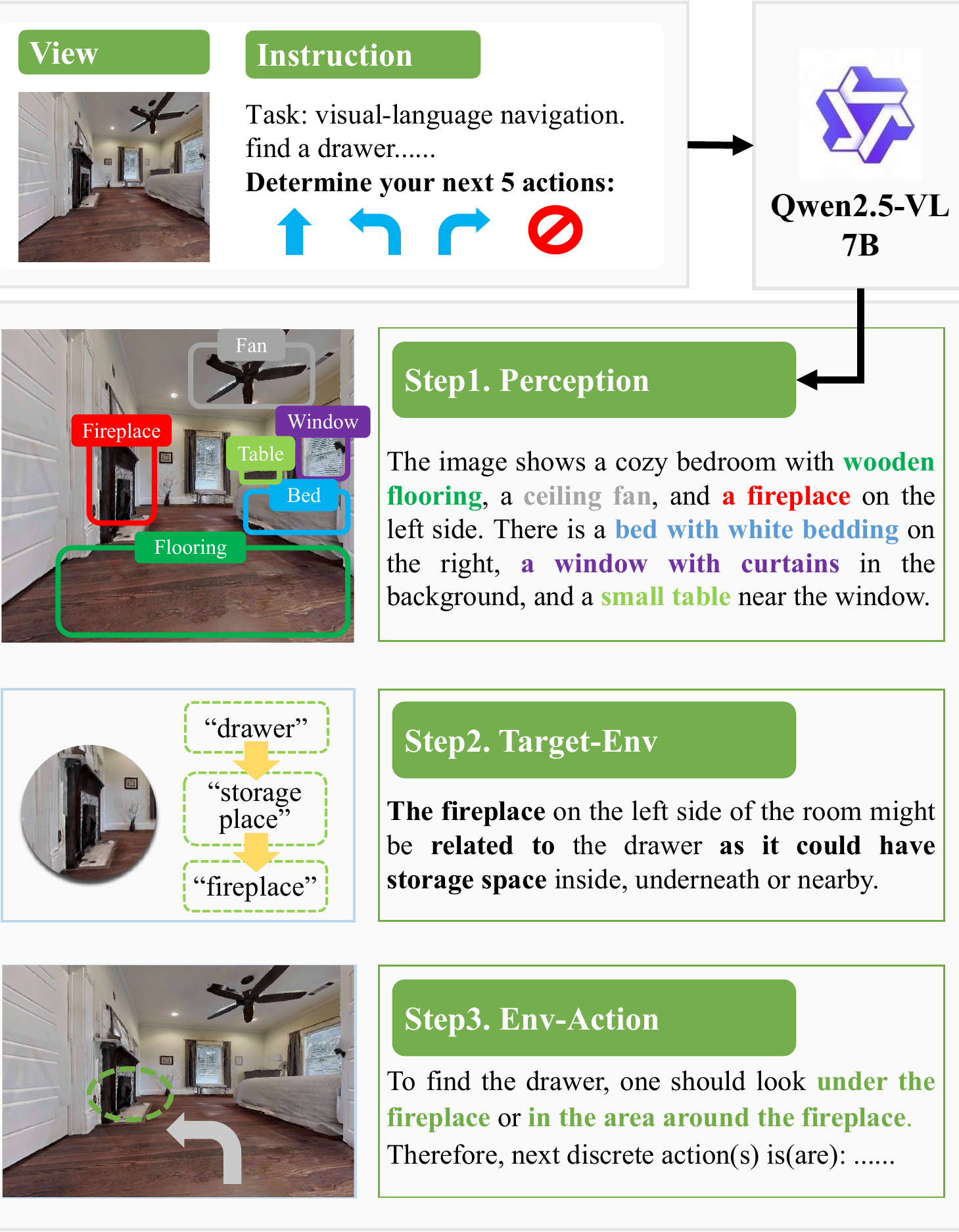}
    \caption{An example showing how our CoT dataset is constructed and what the joint-thinking procedure is upon receiving a new front-view image. }
    \label{fig:CoT_generation_and_joint_thinking_procedure}
\end{figure}


\noindent\textbf{Trajectories and Samples Generation.} We construct object-goal expert trajectories according to task instances in the HM3D-OVON dataset. From each full trajectory, we randomly select one key frame as the current frame, treat preceding frames as history, and take the subsequent five actions as ground-truth. We then combine the instruction text with the rendered images to form the final training samples.

\noindent\textbf{CoT Annotation.} Following the two relationships in $R^2$ mentioned in \cref{sec:intro}, we use Qwen2.5-VL-7B to generate CoT annotations, as illustrated in \cref{fig:CoT_generation_and_joint_thinking_procedure}. Guided by our carefully designed prompt, the structured reasoning should cover three aspects: 1) Environment Perception: a description of the environment based on observations (objects, layout, obstacles, orientation); 2) Target–Environment Relationship: a list of objects and elements potentially related to the target; 3) Environment–Action Relationship: a step-by-step action plan and its justification.

\noindent\textbf{CoT Curation and Refinement.} Due to LLM hallucinations, the raw CoT outputs contain irregularities and errors. We therefore apply a simple cleaning procedure similar to \citep{xiang2025advancing}. Specifically, we discard entries whose “Environment Perception” section contains hedging terms such as “indicate,” “might,” “may,” “imply,” and similar expressions, as uncertain speculation and hallucinations are harmful to downstream object–environment reasoning. Our curated open-vocabulary object-goal navigation dataset contains approximately 300K CoT-annotated samples.

\subsection{\name}
\label{subsec:\name}
\begin{figure*}[ht]
    \centering
    \includegraphics[width=1\textwidth, 
    trim=0.5in 0in 1.3in 0in, clip
    ]{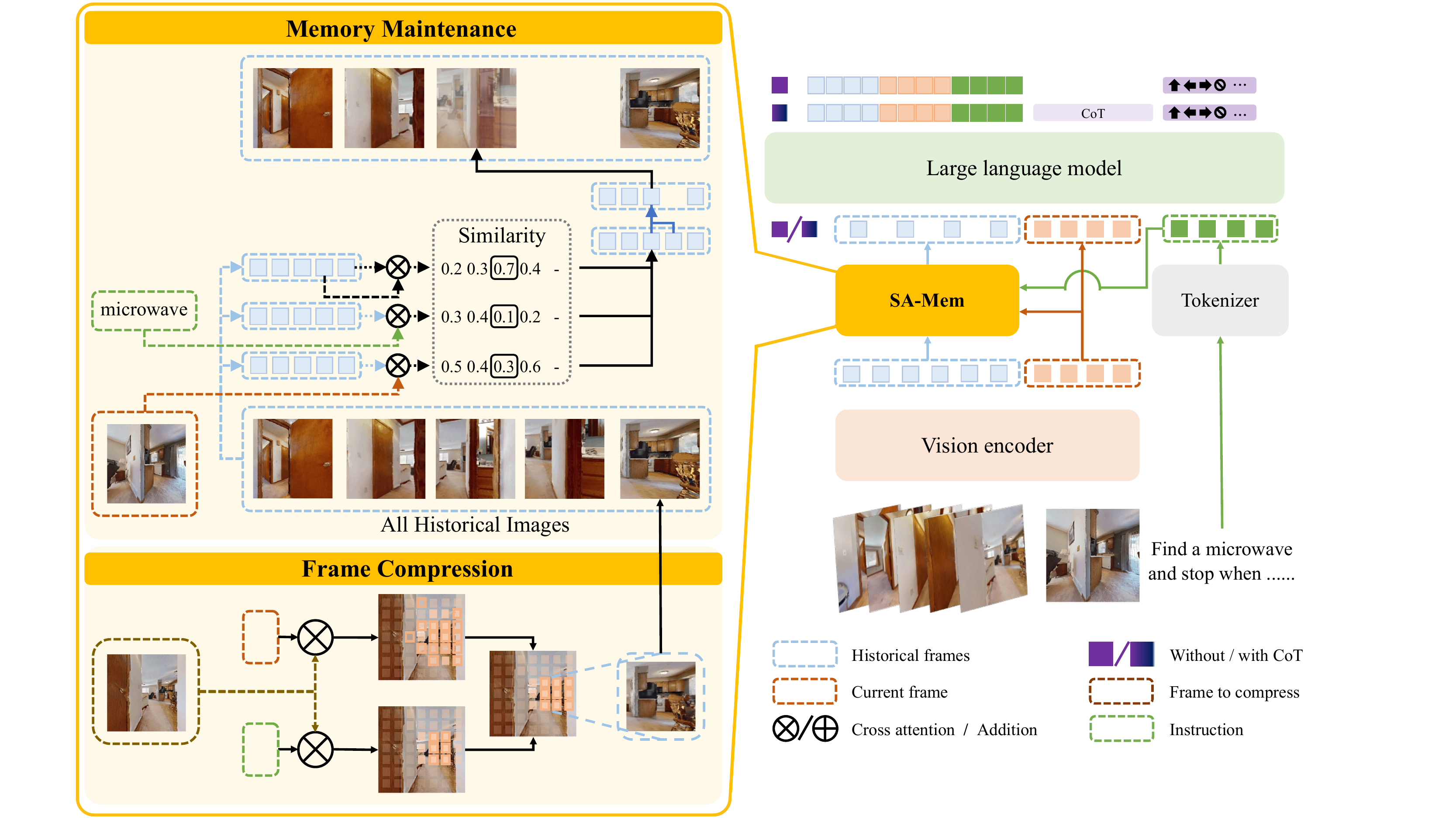}
    \caption{Architecture of \name. Right side shows the architecture, input sequence and output sequence. Left side shows both two mechanisms applied in \mname: 1) Memory Maintenance, and 2) Frame Compression. Mechanisms assist \mname\ to extract and maintain spatially and temporally vital features.}
    \label{fig:pipeline}
\end{figure*}

\noindent\textbf{Pipeline.~} As shown in \cref{fig:pipeline}, our \name~consists of an LLM backbone, a vision encoder and a memory named \mname. The input sequence for the LLM backbone is composed of system prompts, background and task declaration, \mname(including historical frames and current frame) and searching instruction. After vision encoder, \mname\ first compress the frame at the tail if there exists frames in it. Then, \mname\ fuses some pair of adjacent frames dynamically if the length of \mname\ reaches the pre-declared limit. Next, the uncompressed current frame will be appended to \mname. Finally, feed the input sequence to the LLM backbone and the output sequence containing CoT and discrete actions is generated in an auto-regressive way.

\noindent\textbf{\mname.~} 
\mname\ effectively manages both historical and current frames through two effective mechanisms \textbf{1) Frame Compression}, and \textbf{2) Memory Maintenance}. Both of the two mechanisms are illustrated in \cref{alg:memory_management_B4} and \cref{fig:pipeline}.

\noindent\textbf{1) Frame Compression.} 
Specifically, assume $\mathbf{e}_{t\_cur-1}$ is the uncompressed frame embedding at previous timestamp $t\_cur-1$ and is compressed to reduce the number of tokens to 30 with highest similarities through two kinds of cosine similarities upon a new coming frame $\mathbf{e}_{t\_cur}$ as following:
\setlength{\abovedisplayskip}{3pt}
\setlength{\belowdisplayskip}{3pt}
\begin{gather}
\mathcal{S}_{inst_t} = \operatorname{sim}_{\cos}\!\left( \mathbf{e}_{inst}, \mathbf{e}_{t_{\mathrm{cur}-1}} \right)
\end{gather}
\begin{gather}
\mathcal{S}_{cur_t} = \operatorname{sim}_{\cos}\!\left( \mathbf{e}_{t_{\mathrm{cur}}}, \mathbf{e}_{t_{\mathrm{cur}-1}} \right)
\end{gather}
\begin{gather}
\label{gather:s_final}
\mathcal{S}_{final} = \frac{\mathcal{S}_{inst_t} + \mathcal{S}_{cur_t}}{2}
\end{gather}
\begin{gather}
\label{gather:compress_using_topk}
\mathbf{e}_{comp} = \operatorname{TopK}\left(\mathcal{S}_{final}, 30, \mathbf{e}_{t_{\mathrm{cur}-1}}\right)
\end{gather}
where $\mathcal{S}_{inst\_t}$ means the similarity between current instruction $\mathbf{e}_{inst}$ and $\mathbf{e}_{t\_cur-1}$ and $\mathcal{S}_{cur\_t}$ means the similarity between $\mathbf{e}_{t\_cur}$ and $\mathbf{e}_{t\_cur-1}$. This ensures that only the tokens most relevant to the current frame and the target are retained in $\mathbf{e}_{comp}$, thereby improving memory efficiency.

\noindent\textbf{2) Memory Maintenance.} We compute three types of similarities to determine frame fusion as follows:
\begin{gather}
\mathcal{S}_{inst_t} = \mathrm{sim}_{\cos}\!\left( \mathbf{e}_{inst}, \mathbf{e}_{t} \right) \\
\mathcal{S}_{cur_t}  = \mathrm{sim}_{\cos}\!\left( \mathbf{e}_{cur}, \mathbf{e}_{t} \right) \\
\mathcal{S}_{adja_t} = \mathrm{sim}_{\cos}\!\left( \mathbf{e}_{t}, \mathbf{e}_{t+1} \right)
\end{gather}
where $\mathbf{e}_{t}$ and $\mathbf{e}_{t+1}$ are embeddings of two adjacent historical frames at timestamp $t$ and $t+1$, $\mathcal{S}_{inst\_t}$ measures the similarity between the current task instruction embedding $\mathbf{e}_{inst}$ and $\mathbf{e}_{t}$; $\mathcal{S}_{cur\_t}$ measures the similarity between the current frame embedding $\mathbf{e}_{cur}$ and $\mathbf{e}_{t}$; $\mathcal{S}_{adja\_t}$ measures the similarity between adjacent frame embeddings $\mathbf{e}_{t}$ and $\mathbf{e}_{t+1}$ in \mname. We then compute a weighted fusion probability:
\begin{gather}
\mathcal{P}_{inst} = 1 - \mathcal{S}_{inst\_t} \\
\mathcal{P}_{cur} = 1 - \mathcal{S}_{cur\_t} \\
\mathcal{P}_{adja} = \mathcal{S}_{adja\_t} \\
\mathcal{P}_{fuse} = w_{1}\cdot\mathcal{P}_{inst}
+ w_{2}\cdot\mathcal{P}_{cur} 
+ w_{3}\mathcal{P}_{adja}
\end{gather}
The frame pair with the highest fusion probability is identified right now assuming that is at timestamp $t1$, and then their embeddings $\mathbf{e}_{t1}$ and $\mathbf{e}_{t1+1}$ are fused via average pooling to obtain the fused frame representation $\mathbf{e}_{fused\_t1}$:
\begin{gather}
\mathbf{e}_{fused\_t1} = \mathrm{avg\_pool}\!\left( \mathbf{e}_{t1}, \mathbf{e}_{t1+1} \right)
\end{gather}

\begin{algorithm}[t]
\caption{Frame compression and memory maintenance strategies adopted in \mname.}
\label{alg:memory_management_B4} 
\begin{algorithmic}[1]
\Require Current image embeddings $\mathbf{E}_{cur}$, mission text embeddings $\mathbf{T}$ (optional)
\Require Max memory frames $M_{max}$, target tokens $K$
\Ensure Updated memory list $\mathcal{M}$

\Statex {\textbf{Step 1: Compress last frame}}
\If{$|\mathcal{M}| > 0$}
    \State $\mathcal{M}[-1] \gets$ \Call{Compress}{$\mathcal{M}[-1]$, $\mathbf{T}$, $\mathbf{E}_{cur}$}
\EndIf

\Statex {\textbf{Step 2: Merge if exceeds limit}}
\If{$|\mathcal{M}| > M_{max}$}
    \State $\mathbf{w} \gets \{w_{txt}:0.3, w_{adj}:0.4, w_{cur}:0.3\}$
    \State $idx_{mrg} \gets$ \Call{FindMergePair}{$\mathbf{E}_{cur}$, $\mathbf{T}$}
    \State \Call{Merge}{$idx_{mrg}$, $K$}
\EndIf

\Statex {\textbf{Step 3: Add current frame}}
\State $\mathcal{M}.\text{append}(\mathbf{E}_{cur})$

\Statex

\Function{FindMergePair}{$\mathbf{E}_{cur}$, $\mathbf{T}$}
    \State $\bar{\mathbf{t}} \gets \text{mean}(\mathbf{T})$ if $\mathbf{T} \neq \text{None}$
    \State $\bar{\mathbf{c}} \gets \text{mean}(\mathbf{E}_{cur})$
    \State $\mathbf{s} \gets []$
    \For{$i = 0$ to $|\mathcal{M}| - 2$}
        \State $\bar{\mathbf{a}} \gets \text{mean}(\mathcal{M}[i])$, $\bar{\mathbf{b}} \gets \text{mean}(\mathcal{M}[i+1])$
        \State $p \gets 0$
        \If{$w_{adj} > 0$}
            \State $p \gets p + w_{adj} \cdot \cos(\bar{\mathbf{a}}, \bar{\mathbf{b}})$
        \EndIf
        \If{$w_{txt} > 0$ and $\mathbf{T} \neq \text{None}$}
            \State $p \gets p + w_{txt} \cdot \left(1 - \frac{\cos(\bar{\mathbf{a}}, \bar{\mathbf{t}}) + \cos(\bar{\mathbf{b}}, \bar{\mathbf{t}})}{2}\right)$
        \EndIf
        \If{$w_{cur} > 0$}
            \State $p \gets p + w_{cur} \cdot \left(1 - \frac{\cos(\bar{\mathbf{a}}, \bar{\mathbf{c}}) + \cos(\bar{\mathbf{b}}, \bar{\mathbf{c}})}{2}\right)$
        \EndIf
        \State $\mathbf{s}.\text{append}(p)$
    \EndFor
    \State \Return $\arg\max(\mathbf{s})$
\EndFunction

\Statex

\Function{Merge}{$idx$, $K$}
    \State $\mathbf{E}_{mrg} \gets \text{concat}(\mathcal{M}[idx], \mathcal{M}[idx+1])$
    \State $\mathbf{E}_{pool} \gets \text{AvgPool1d}(\mathbf{E}_{mrg}, K)$
    \State $\mathcal{M}[idx] \gets \mathbf{E}_{pool}$
    \State $\mathcal{M}.\text{pop}(idx+1)$, $\mathcal{I}.\text{pop}(idx+1)$
\EndFunction

\end{algorithmic}
\end{algorithm}

\section{Experiments}

\subsection{Implementation Details}

\noindent\textbf{Training.~} Our \name~ is based on Qwen2.5-VL-7B. The hyperparameters of training are listed in \cref{tab:implementation_details}. All the input images will be resized to $640 \times 520$ for easy coding. It is important that our \name~ only receives the first-person view RGB image and is only trained through SFT for $1$ epoch without other visual modalities or auxiliary modules like dynamically constructed map. The training costs around $1280$ gpu hours in total on $64$ $H20$ gpus.
\begin{table}[!t]
    \setlength{\belowcaptionskip}{-5pt}
    \setlength{\abovecaptionskip}{5pt}
\caption{Hyperparameters and settings for both training and evaluation stages. Terms \textit{LR}, \textit{BS}, \textit{Img}, \textit{Success Dist}, \textit{Cam}, \textit{Inp Mod}, \textit{F-P}, \textit{R2xR}
and \textit{S.RGB}
are abbreviations for \textit{deepspeed}, \textit{learning rate}, \textit{batch size}, \textit{image}, \textit{success distance}, \textit{camera}, \textit{input modality}, \textit{First-Person}, \textit{both R2R and RXR datasets}
and \textit{Single RGB}. }
\label{tab:implementation_details}
\centering
\small
\resizebox{0.48\textwidth}{!}{
    \begin{tabular}{
        >{\centering\arraybackslash}m{0.2\linewidth}
        >{\centering\arraybackslash}m{0.3\linewidth}
        >{\centering\arraybackslash}m{0.3\linewidth}
        >{\centering\arraybackslash}m{0.2\linewidth}
    }
        \toprule
        \multicolumn{2}{c}{\textbf{Training}} &
        \multicolumn{2}{c}{\textbf{Evaluation}} \\
        \textbf{Parameters} & \textbf{Value} & \textbf{Parameters} & \textbf{Value} \\
        \midrule
        Inp Mod & S.RGB & Inp Mod & S.RGB \\
        Views & F-P & Views & F-P \\
        Warmup & 0.05 & Success Dist & 1\,m \\
        Datasets & R2xR; OVON~\citep{yokoyama2024hm3d@hm3dovon} & Turn Angle & 30$^{\circ}$ \\
        Activated & Aligner \& LLM & Step Size & $0.25\,m$ \\
        Epoch & 1 & Cam Width & 720 \\
        Total BS & 256 & Cam Height & 640 \\
        Dtype & bfloat16 & Cam Position & 0.88\,m \\
        LR & $2e$$^{-4}$ & Cam. HFOV & 110$^{\circ}$ \\
        Img W & 640 & Img W & 640 \\
        Img H & 520 & Img H & 520 \\
        \bottomrule
    \end{tabular}
}
\vspace{-6mm}
\end{table}

\noindent\textbf{Evaluation.~} 
Seen from \cref{tab:implementation_details}, our \name\ receives and processes a single first-person RGB frame at each step during evaluation. \textit{Success distance} means if agent is only 1\,meter away from the target then the task is accomplished, \textit{turn angle} and \textit{step size} specify the extent to which the agent can move and turn in each action. Camera related parameters, \textit{Width}, \textit{Height}, \textit{Position}, and \textit{HFOV} specify the size of the input images, the height of the camera position and the extent of the observable scene captured by a camera in the horizontal direction. Lastly, we prepend \textit{Please do not output your thinking process.} to the prompt template to drive \name\ to output without CoT, achieving 2Hz inference.

\subsection{Main Results on HM3D-OVON}

\begin{table*}[!t]  
    \centering
    \setlength{\belowcaptionskip}{-5pt}  
    \setlength{\abovecaptionskip}{5pt} 
    \caption{Main results on open-vocabulary object-goal navigation task and OVON~\citep{yokoyama2024hm3d@hm3dovon} dataset. We set the background color to light yellow and blue to indicate the most comparable method to \name\ without depth map and the methods with depth map. Comparing to Uni-Navid\citep{zhang2024uni@uninavid}, our \name\ achieves the highest SR on all val-datasets without depth data. The sub-columns, \textit{SFT}, \textit{RL}, \textit{S.RGB}, \textit{Pano}, \textit{Dep}, \textit{Odo}, \textit{Sim}, \textit{QA} are abbreviations for \textit{supervied fine-tuning}, \textit{re-inforcement learning}, \textit{single-rgb}, \textit{panoptic}, \textit{depth}, \textit{odometry}, \textit{simulator}, \textit{question answering} respectively. }
    \resizebox{\textwidth}{!}{
        \begin{tabular}{c|ccc|ccccc|c|cc|cc|cc}
            \toprule[1.1pt]
            \textbf{Method} & \multicolumn{3}{c|}{\textbf{Model Inputs}} & \multicolumn{5}{c|}{\textbf{Training Setting}} & \textbf{Map} & \multicolumn{2}{c|}{\textbf{Val-Seen}} & \multicolumn{2}{c|}{\textbf{Val-Seen-Synonyms}} & \multicolumn{2}{c}{\textbf{Val-Unseen}} \\
            
             & S.RGB & Dep & Odo & SFT & RL & Sim & Real & QA &  & \textbf{SR}$\uparrow$ & \textbf{SPL}$\uparrow$ & \textbf{SR}$\uparrow$ & \textbf{SPL}$\uparrow$ & \textbf{SR}$\uparrow$ & \textbf{SPL}$\uparrow$ \\
            \midrule[0.7pt]

            BC~\citep{pomerleau1988alvinn@alvinn}      & \Checkmark & & &  & & \Checkmark &  &  & & 11.1 & 4.5  & 9.9  & 3.8  & 5.4  & 1.9  \\
            DAgger~\citep{ross2011reduction}      & \Checkmark & & &  & & \Checkmark &  &  &  & 11.1 & 4.5  & 9.9  & 3.8  & 5.4  & 1.9 \\
            RL~\citep{schulman2017proximal}      & \Checkmark & & & & \Checkmark & \Checkmark &  &  &   & 18.1 & 9.4 & 15.0 & 7.4 & 10.2 & 4.7 \\
            DAgRL~\citep{chen2019touchdown@touchdown}   & \Checkmark & & & \Checkmark & \Checkmark & \Checkmark &  &  &  & 41.3 & 21.2 & 29.4 & 14.4 & 18.3 & 7.9  \\
            BCRL~\citep{wang2019reinforced}   & \Checkmark & & & \Checkmark & \Checkmark & \Checkmark &  &  & & 39.2 & 18.7 & 27.8 & 11.7 & 18.6 & 7.5  \\

            VLFM~\citep{yokoyama2024vlfm}    & \Checkmark & \Checkmark &  \Checkmark &  & \Checkmark & \Checkmark &  &  & \Checkmark & 35.2 & 18.6 & 32.4 & 17.3 & 35.2 & 19.6 \\
            DAgRL+OD~\citep{yokoyama2024hm3d@hm3dovon}  & \Checkmark & & & \Checkmark & \Checkmark & \Checkmark &  &  & & 38.5 & 21.1 & 39.0 & 21.4 & 37.1 & 19.8 \\
        
            \rowcolor{methodgroup}
            Nav-R1~\citep{liu2025nav@navr1} & \Checkmark & \Checkmark &  & \Checkmark & \Checkmark & \Checkmark &  & \Checkmark &  & 58.4 & 26.3 & 48.1 & 23.1 & 42.2 & 20.1 \\
            \rowcolor{methodgroup}
            MTU3D~\citep{zhu2025move@mtu3d}  & \Checkmark & \Checkmark &  & \Checkmark &  & \Checkmark & \Checkmark & \Checkmark & \Checkmark &  55.0 & 23.6 & 45.0 & 14.7 & 40.8 & 12.1 \\
            
            \rowcolor{yellow!20}
            Uni-NaVid~\citep{zhang2024uni@uninavid}  & \Checkmark & &  & \Checkmark &  & \Checkmark & \Checkmark & \Checkmark &  & 41.3 & 21.1 & 43.9 & 21.8 & 39.5 & 19.8 \\
            \midrule[0.7pt]
            \textbf{\name} & \Checkmark  & &  & \Checkmark  & &  \Checkmark &  & &  & 45.6 & 21.0 & 45.9 & 21.1 & 44.0 & 18.0 \\
            \bottomrule[1.1pt]
        \end{tabular}
        \label{tab:main_results_on_open_vocabulary_object_goal_navigation_task_and_OVON}
    }
\end{table*}

\cref{tab:main_results_on_open_vocabulary_object_goal_navigation_task_and_OVON} compares the performance of \name\ against existing approaches on the HM3D-OVON dataset~\citep{yokoyama2024hm3d@hm3dovon}. Notably, \name\ operates using \textbf{only first-person RGB observations} and is trained exclusively on simulated data via supervised learning. As depth maps provide explicit geometric cues that directly encode 3D spatial relationships, obstacle distances, and traversable space boundaries—information that must be implicitly inferred from monocular RGB observations, which alters the nature of the navigation task and results in model overfitting training datasets, we compare \name\ with others in two dimensions: methods with depth map and that without depth map.

\noindent\textbf{Methods without Depth Map.~}
Among prior works listed in \cref{tab:main_results_on_open_vocabulary_object_goal_navigation_task_and_OVON}, the methods most comparable to \name\ is Uni-Navid\citep{zhang2024uni@uninavid}. On all the \emph{val-seen}, \emph{val-seen-synonyms}, and \emph{val-unseen} splits, \name\ successfully achieves higher Success Rate (SR), while maintaining comparable Success weighted by Path Length (SPL). The slightly lower SPL values compared to can be attributed to its significantly larger and more diverse training corpus, which includes question-answering datasets, large-scale real-world driving and navigation data, and diverse multimodal signals. Real-world data provides richer spatial layouts and object distributions, enhancing search efficiency, while question-answering datasets improve inherent language modeling capabilities of its underlying LLM. Lastly, the metric SPL in object-goal navigation tasks could be not as effective as that in instrct-goal navigation tasks as there could be more than one target objects distributed in the indoor environments and paths to targets differs a lot, but the ground-truth path used for SPL metric is the same all the time.

\noindent\textbf{Methods with Depth Map.~}
For other methods requiring map such as Nav-R1\citep{liu2025nav@navr1} and MTU3D\citep{zhu2025move@mtu3d}, we observe that both achieve stronger SR and SPL on the \emph{val-seen} split, which corresponds to a closed-set setting. 
We believe that this is due to the extra depth map input and much more kinds of dataset resulting the trained model to be overfitted to the seen objects.
In contrast, \name\ delivers competitive SR on \emph{val-seen-synonyms}, surpasses MTU3D on both SR and SPL for \emph{val-seen-synonyms} and \emph{val-unseen}, and outperforms Nav-R1 on \emph{val-unseen} across both metrics. We attribute these gains to our explicit and joint relational modeling—capturing target-environment and environment–action relationships—combined with our \mname, which consistently extracts and maintains target- and current observation-relevant features throughout navigation. These components enable \name\ to generalize effectively in open-vocabulary settings.

\subsection{Ablation Study}
\label{subsection:ablation_study}

This section mainly introduce the experiments we conducted to support effectiveness of the CoT dataset construction and our \mname. For CoT dataset construction, it remains to be validated that what's the difference before and after applying CoT dataset during training, and that what's the significance of each component in the CoT. While for \mname, we illustrate in this section the mechanisms including frame compression and memory maintenance.

\noindent\textbf{Effection of CoT~}
As shown by the first line in \cref{tab:ablation_results_of_components_in_CoT}, after tuning \name\ without \mname\ for one epoch on OVON\citep{yokoyama2024hm3d@hm3dovon}, the final model gets only 14.8 and 10.0 for SR and SPL metrics respectively. However, seen from the forth line, applying only CoT enhances the model and increases the performance to 28.4 and 17.1 for metrics SR and SPL respectively on val-unseen dataset. Besides, results on val-seen and val-seen-synonyms are also far surpass that without CoT. This improvements proves that the CoT, modeling both relations explicitly, is indeed effective and essential for open-vocabulary object-goal navigation task.

\noindent\textbf{Components of CoT.~} 
As shown in \cref{tab:ablation_results_of_components_in_CoT}, components in CoT matter when training and evaluating our \name. It can be seen that when the CoT simultaneously includes perception results, Target-Env relationship reasoning, and Env–Task relationship reasoning, the model achieves the highest “SR” and “SPL” scores on all three datasets: val-seen, val-seen-synonyms, and val-unseen. Furthermore, as shown in the 2nd and 4th rows of the table, explicitly reasoning about both the Target-Env and Env–Task relationships yields improvements of 6.8, 2.7, and 3.9 in SR on the three datasets, respectively, demonstrating the effectiveness of jointly reasoning about the two relationships.

\begin{table}[t]
    \centering
    \setlength{\belowcaptionskip}{-5pt}  
    \setlength{\abovecaptionskip}{5pt} 
    \caption{
        Ablation results of each components in CoT. Experiments are conducted on \name\ without \mname\ to avoid their potential effection.
        \textit{Percep}, \textit{Target-Env} and \textit{Env-Task} are used as abbreviations for \textit{Perception}, \textit{Target-Environment Relationship} and \textit{Environment-Task Relationship} respectively. \textit{Percep} indicates the environment perception results.}
    \resizebox{0.45\textwidth}{!}{   
        \begin{tabular}{ccc|cc|cc|cc}
            \toprule[1.1pt]
            \multicolumn{3}{c|}{\textbf{Components}} & \multicolumn{2}{c|}{\textbf{Val-Seen}} & \multicolumn{2}{c|}{\textbf{Val-Seen-Synonyms}} & \multicolumn{2}{c}{\textbf{Val-Unseen}} \\
            Percep & Target-Env & Env-Task &  \textbf{SR}$\uparrow$ & \textbf{SPL}$\uparrow$ & \textbf{SR}$\uparrow$ & \textbf{SPL}$\uparrow$ & \textbf{SR}$\uparrow$ & \textbf{SPL}$\uparrow$ \\

            \midrule[0.7pt]
             & & & 22.7 & 14.8 & 17.4 & 11.8 & 14.8 & 10.0 \\

             \Checkmark & & & 25.4 & 16.5 & 28.1 & 17.2 & 24.5 & 15.9 \\
             \Checkmark & \Checkmark & & 29.1 & 18.0 & 27.8 & 17.9 & 25.4 & 16.3 \\
             \Checkmark & \Checkmark & \Checkmark & 32.2 & 18.8 & 30.8 & 18.8 & 28.4 & 17.1 \\

            \bottomrule[1.1pt]
        \end{tabular}
    }
    \label{tab:ablation_results_of_components_in_CoT}
\end{table}

\noindent\textbf{Frame Compression Strategies.~} We compare in total two types of compressing strategies. One strategy is compressing with both instructions and current frame features mentioned in \cref{fig:pipeline}, while the other one is compressing with instructions only and the key difference compared with \cref{gather:s_final} lies in computing process of $S_{final}$ in \cref{gather:s_final}:
\begin{gather}
\label{gather:s_final_instruction_only}
\mathcal{S}_{final} = \mathcal{S}_{inst\_t}
\end{gather}

Then, 30 embeddings with highest similarity are selected with reference to \cref{gather:compress_using_topk}.
In a word, the comparation in \cref{tab:ablation_results_of_strategies_to_compress_a_frame_to_be_a_historical_frame} shows that considering both current frame and current instruction is the best solution to compress a frame, as it is able to extract task-related and current observation-related key information from the historical frames to avoid information redundancy.

\begin{table}[t]
    \centering
    \setlength{\belowcaptionskip}{-5pt}  
    \setlength{\abovecaptionskip}{5pt} 
    \caption{Ablate the strategies to compress a frame to be a historical frame. Experiments are conducted on \name\ with only the \mname\ modified.}
    \resizebox{0.45\textwidth}{!}{   
        \begin{tabular}{cc|cc|cc|cc}
            \toprule[1.1pt]
            \multicolumn{2}{c|}{\textbf{Strategies}} & \multicolumn{2}{c|}{\textbf{Val-Seen}} & \multicolumn{2}{c|}{\textbf{Val-Seen-Synonyms}} & \multicolumn{2}{c}{\textbf{Val-Unseen}} \\
            Instruction & Current Frame &  \textbf{SR}$\uparrow$ & \textbf{SPL}$\uparrow$ & \textbf{SR}$\uparrow$ & \textbf{SPL}$\uparrow$ & \textbf{SR}$\uparrow$ & \textbf{SPL}$\uparrow$ \\

            \midrule[0.7pt]
             \Checkmark &  & 42.2 & 21.5 & 37.5 & 20.6 & 39.8 & 20.5 \\
             \Checkmark & \Checkmark & 45.0 & 21.1 & 43.2 & 20.9  & 42.0 & 18.8 \\
             
            \bottomrule[1.1pt]
        \end{tabular}
    }
    \label{tab:ablation_results_of_strategies_to_compress_a_frame_to_be_a_historical_frame}
\end{table}

\noindent\textbf{Memory Maintenance Strategies.~}
Intuitively, When the memory buffer reaches its maximum capacity, a maintenance mechanism is triggered to selectively remove or fuse frames while preserving the most informative historical context. As simply discarding the earliest frame from the historical sequence may lead to substantial loss of critical early contextual information, ablation studies are conducted to evaluate two operations: \textbf{1) removing a single frame}; and \textbf{2) fusing a pair of adjacent frames into one} as shown in the first column in \cref{tab:ablation_results_of_strategies_to_maintain_the_memory_when_reaches_the_maximum_length}. A critical aspect common to both operations is the criterion for selecting the frame(s) to be removed or fused. We investigate two distinct criteria, \textbf{\textit{1) Temporal Interval-Based}} and \textbf{\textit{2) Relevance-Based}}, for each operation. The \textit{Temporal Interval-Based} criterion is shown in \cref{alg:memory_management_B1}, maintaining temporal diversity by removing frames that are temporally redundant. The \textit{Relevance-Based} criterion is shown in \cref{alg:memory_management_B4}. 

Ablation results are shown in \cref{tab:ablation_results_of_strategies_to_maintain_the_memory_when_reaches_the_maximum_length}, where applying both fusion operation and relavance-based criterion at the same time achieves the highest performance. It is evident that the fusion operation and relavance-based criterion clearly outperforms others. We posit that this is because frame merging facilitates the preservation of similar key visual information present in the original frames, while enabling uncorrelated noise components to counteract each other. Consequently, the merged features present more salient visual-semantic information and reduced noise levels.

\begin{algorithm}[t]
\caption{Removal operation with the temporal interval-based criterion.}
\label{alg:memory_management_B1}
\begin{algorithmic}[1]
\Require RGB images $\mathcal{R} = \{rgb_1, rgb_2, \ldots, rgb_n\}$, camera pose $p$
\Require Maximum history frames $M_{hist}$, current frames $N_{curr}$
\Require Input image size $(W, H)$, history resize ratio $\alpha$
\Ensure Updated frame lists: $\mathcal{L}_{rgb}$, $\mathcal{L}_{pose}$, $\mathcal{L}_{idx}$

\State \textbf{Add New Frames}
\State $\mathcal{R}_{new} \gets \emptyset$

\State $\mathcal{L}_{rgb} \gets \mathcal{L}_{rgb} \cup \mathcal{R}_{new}$
\State $\mathcal{L}_{pose} \gets \mathcal{L}_{pose} \cup \{p\}^{|\mathcal{R}_{new}|}$ 
\State $\mathcal{L}_{idx} \gets \mathcal{L}_{idx} \cup \{t_{global}\}^{|\mathcal{R}_{new}|}$
\State $t_{global} \gets t_{global} + 1$

\State
\State \textbf{Remove Redundant Frame}
\If{$|\mathcal{L}_{rgb}| > M_{hist} + N_{curr}$}
    \State $\mathcal{L}_{idx}^{hist} \gets \mathcal{L}_{idx}[0 : -N_{curr}]$
    \State $\Delta \gets [\mathcal{L}_{idx}^{hist}[i+1] - \mathcal{L}_{idx}^{hist}[i]]_{i=0}^{|\mathcal{L}_{idx}^{hist}|-2}$
    \State $idx_{min} \gets \arg\min_{i} \Delta[i]$
    \State $idx_{rv} \gets idx_{min} + 1$
    \State Remove $\mathcal{L}_{rgb}[idx_{rv}]$, $\mathcal{L}_{pose}[idx_{rv}]$, $\mathcal{L}_{idx}[idx_{rv}]$
\EndIf

\State \Return $\mathcal{L}_{rgb}$, $\mathcal{L}_{pose}$, $\mathcal{L}_{idx}$
\end{algorithmic}
\end{algorithm}

\begin{table}[t]
    \centering
    \setlength{\belowcaptionskip}{-5pt}  
    \setlength{\abovecaptionskip}{5pt} 
    \caption{Ablation results of strategies to maintain the memory when reaches the maximum length. Experiments are conducted on \name\ with CoT, only modifying \mname.}
    \resizebox{0.45\textwidth}{!}{   
        \begin{tabular}{cccc|cc|cc|cc}
            \toprule[1.1pt]
            
            \multicolumn{2}{c}{\textbf{Operations}} & \multicolumn{2}{c|}{\textbf{Criteria}} & \multicolumn{2}{c|}{\textbf{Val-Seen}} & \multicolumn{2}{c|}{\textbf{Val-Seen-Synonyms}} & \multicolumn{2}{c}{\textbf{Val-Unseen}} \\
            
            Removal & Fusion & Temp & Rele & \textbf{SR}$\uparrow$ & \textbf{SPL}$\uparrow$ & \textbf{SR}$\uparrow$ & \textbf{SPL}$\uparrow$ & \textbf{SR}$\uparrow$ & \textbf{SPL}$\uparrow$ \\

            \midrule[0.7pt]

             \Checkmark & & \Checkmark & & 45.0 & 21.1 & 43.2 & 20.9  & 42.0 & 18.8 \\ 
             \Checkmark & & & \Checkmark & 47.7 & 20.6 & 44.8 & 20.5  & 41.1 & 16.4 \\ 
             & \Checkmark & \Checkmark  & &  43.4 & 21.8 & 43.1 & 21.8 & 39.5 & 20.2 \\
             & \Checkmark & & \Checkmark & 45.6 & 21.0 & 45.9 & 21.1 & 44.0 & 18.0 \\ 
            \bottomrule[1.1pt]
        \end{tabular}
    }
    \label{tab:ablation_results_of_strategies_to_maintain_the_memory_when_reaches_the_maximum_length}
\end{table}

\begin{figure*}[!t]
    \centering
    \includegraphics[width=0.95\textwidth, 
    trim=0.1in 0in 0.1in 0in, clip
    ]{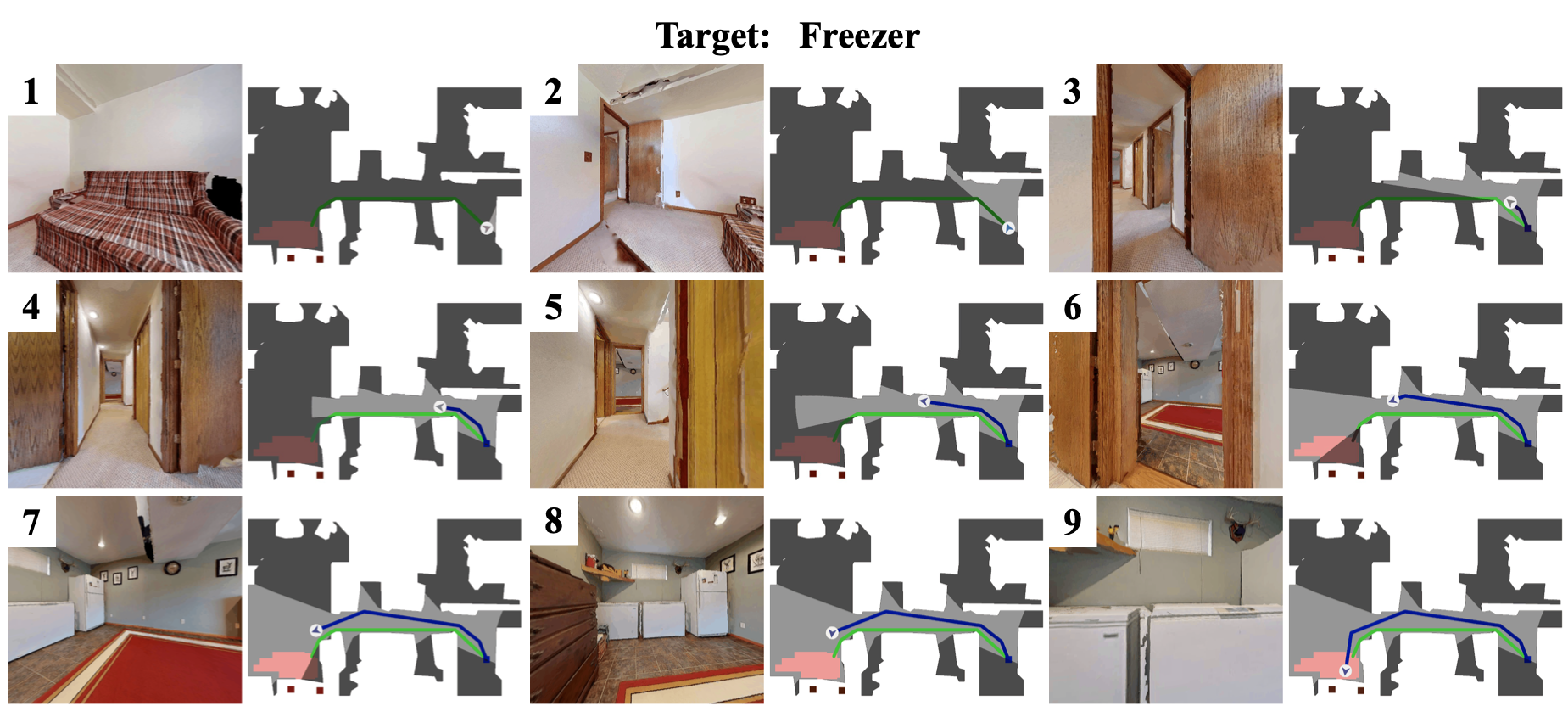}
    \caption{Visualization of trajectory on OVON\citep{yokoyama2024hm3d@hm3dovon} val-unseen dataset during evaluation. There are in total tens of frames and we select and show the key frames here.}
    \label{fig:unseen}
\end{figure*}

\subsection{Qualitative Results}
\noindent\textbf{Visualization of trajectory searching unseen objects.}
We save the whole trajectory on OVON\citep{yokoyama2024hm3d@hm3dovon} val-unseen dataset to local files, and show some key frames in \cref{fig:unseen}. Agent is searching for a freezer. Initially, the agent is in a bedroom facing a bed. Then, it turns around and walk out the room heading for living room. Next, when entering the living room it can see the freezer is on the left side. Finally, it turns left and forward to the target freezer. Though at the beginning our \name\ is not in the same room where target object is, \name\ is able to realize walking out of the bedroom, ignore the adjacent bedroom and the stairs, and finally enter the living room successfully approaching the freezr.

\begin{figure}
    \centering
    \includegraphics[width=1\linewidth,
        trim=0.1in 0in 0.1in 0in, clip
    ]{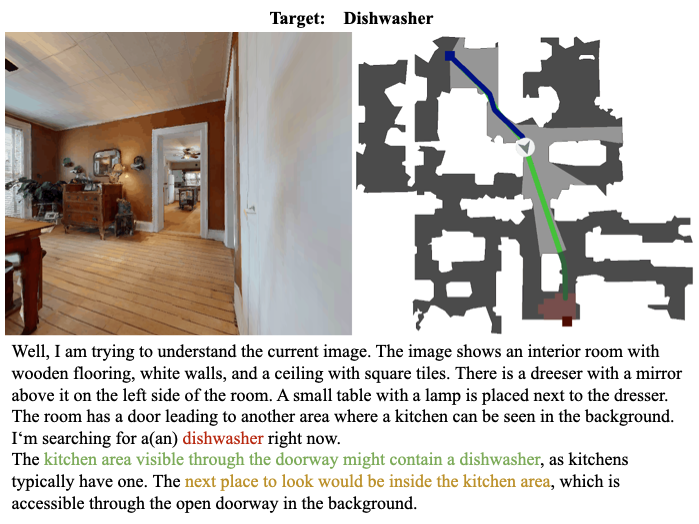}
    \caption{Visualization of CoT content when our \name\ is searching for an unseen object in dataset\citep{yokoyama2024hm3d@hm3dovon}. \name\ first percept the surroundings, then identify and finally make a plan. Blue line is how the agent actually walks, and the green line is the expert trajectory collected and saved in the dataset.}
    \label{fig:1111}
\end{figure}

\noindent\textbf{Visualization of CoT during searching unseen objects.}
Seen from \cref{fig:1111}, \name\ is on its way to the target object, dishwasher. Blue line is how the agent actually walks, and the green line is the expert trajectory collected and saved in the dataset. As the text content shows, \name\ first percept and realize that the kitchen area in the background might contain a dishwasher, and finally make a plan to go through the doorway to enter the kitchen. The way agent walks along is almost the same as the expert trajectory, which shows our strong generalization to searching an unseen object, validating the effectiveness of our framework.

\section{Conclusion}

In this work, we propose a framework emphasizing that explicitly modeling two relationships enhances a navigation model's searching capability, expecially for generalization to unseen objects. Moreover, our \name\ model trained on our own ObjectNav CoT dataset achieves the best SR metric showing outstanding generalization capability and validating the solidness of our framework. Hope the results in this work and the materials we release could help advance the research in ObjectNav tasks.

{
    \small
    \bibliographystyle{ieeenat_fullname}
    \bibliography{main}

@String(CVPR= {IEEE Conf. Comput. Vis. Pattern Recog.})

@String(ICCV= {Int. Conf. Comput. Vis.})

@String(ECCV= {Eur. Conf. Comput. Vis.})

@String(CVPR  = {CVPR})

@String(ICCV  = {ICCV})

@String(ECCV  = {ECCV})

@article{pomerleau1988alvinn@alvinn,
  title={Alvinn: An autonomous land vehicle in a neural network},
  author={Pomerleau, Dean A},
  journal={Advances in neural information processing systems},
  volume={1},
  year={1988}
}

@inproceedings{ross2011reduction,
  title={A reduction of imitation learning and structured prediction to no-regret online learning},
  author={Ross, St{\'e}phane and Gordon, Geoffrey and Bagnell, Drew},
  booktitle={Proceedings of the fourteenth international conference on artificial intelligence and statistics},
  pages={627--635},
  year={2011},
  organization={JMLR Workshop and Conference Proceedings}
}

@inproceedings{chen2019touchdown@touchdown,
  title={Touchdown: Natural language navigation and spatial reasoning in visual street environments},
  author={Chen, Howard and Suhr, Alane and Misra, Dipendra and Snavely, Noah and Artzi, Yoav},
  booktitle={Proceedings of the IEEE/CVF Conference on Computer Vision and Pattern Recognition},
  pages={12538--12547},
  year={2019}
}

@inproceedings{wang2019reinforced,
  title={Reinforced cross-modal matching and self-supervised imitation learning for vision-language navigation},
  author={Wang, Xin and Huang, Qiuyuan and Celikyilmaz, Asli and Gao, Jianfeng and Shen, Dinghan and Wang, Yuan-Fang and Wang, William Yang and Zhang, Lei},
  booktitle={Proceedings of the IEEE/CVF conference on computer vision and pattern recognition},
  pages={6629--6638},
  year={2019}
}

@inproceedings{yokoyama2024hm3d@hm3dovon,
  title={Hm3d-ovon: A dataset and benchmark for open-vocabulary object goal navigation},
  author={Yokoyama, Naoki and Ramrakhya, Ram and Das, Abhishek and Batra, Dhruv and Ha, Sehoon},
  booktitle={2024 IEEE/RSJ International Conference on Intelligent Robots and Systems (IROS)},
  pages={5543--5550},
  year={2024},
  organization={IEEE}
}

@article{zhang2024uni@uninavid,
  title={Uni-navid: A video-based vision-language-action model for unifying embodied navigation tasks},
  author={Zhang, Jiazhao and Wang, Kunyu and Wang, Shaoan and Li, Minghan and Liu, Haoran and Wei, Songlin and Wang, Zhongyuan and Zhang, Zhizheng and Wang, He},
  journal={arXiv preprint arXiv:2412.06224},
  year={2024}
}

@inproceedings{zhu2025move@mtu3d,
  title={Move to understand a 3d scene: Bridging visual grounding and exploration for efficient and versatile embodied navigation},
  author={Zhu, Ziyu and Wang, Xilin and Li, Yixuan and Zhang, Zhuofan and Ma, Xiaojian and Chen, Yixin and Jia, Baoxiong and Liang, Wei and Yu, Qian and Deng, Zhidong and others},
  booktitle={Proceedings of the IEEE/CVF International Conference on Computer Vision},
  pages={8120--8132},
  year={2025}
}

@article{liu2025nav@navr1,
  title={Nav-r1: Reasoning and navigation in embodied scenes},
  author={Liu, Qingxiang and Huang, Ting and Zhang, Zeyu and Tang, Hao},
  journal={arXiv preprint arXiv:2509.10884},
  year={2025}
}

@misc{zhang2025embodiednavigationfoundationmodel@navfom,
      title={Embodied Navigation Foundation Model}, 
      author={Jiazhao Zhang and Anqi Li and Yunpeng Qi and Minghan Li and Jiahang Liu and Shaoan Wang and Haoran Liu and Gengze Zhou and Yuze Wu and Xingxing Li and Yuxin Fan and Wenjun Li and Zhibo Chen and Fei Gao and Qi Wu and Zhizheng Zhang and He Wang},
      year={2025},
      eprint={2509.12129},
      archivePrefix={arXiv},
      primaryClass={cs.RO},
      url={https://arxiv.org/abs/2509.12129}, 
}

@misc{chaplot2020objectgoalnavigationusing,
      title={Object Goal Navigation using Goal-Oriented Semantic Exploration}, 
      author={Devendra Singh Chaplot and Dhiraj Gandhi and Abhinav Gupta and Ruslan Salakhutdinov},
      year={2020},
      eprint={2007.00643},
      archivePrefix={arXiv},
      primaryClass={cs.CV},
      url={https://arxiv.org/abs/2007.00643}, 
}

@inproceedings{xiang2025advancing,
  title={Advancing Visual Large Language Model for Multi-granular Versatile Perception},
  author={Xiang, Wentao and Tan, Haoxian and Zhong, Yujie and Wei, Cong and Li, Dengjie and Yang, Yujiu},
  booktitle={Proceedings of the IEEE/CVF International Conference on Computer Vision},
  pages={22153--22164},
  year={2025}
}

@article{majumdar2022zson,
  title={Zson: Zero-shot object-goal navigation using multimodal goal embeddings},
  author={Majumdar, Arjun and Aggarwal, Gunjan and Devnani, Bhavika and Hoffman, Judy and Batra, Dhruv},
  journal={Advances in Neural Information Processing Systems},
  volume={35},
  pages={32340--32352},
  year={2022}
}

@article{zhang2018semantic,
  title={Semantic SLAM based on object detection and improved octomap},
  author={Zhang, Liang and Wei, Leqi and Shen, Peiyi and Wei, Wei and Zhu, Guangming and Song, Juan},
  journal={IEEE Access},
  volume={6},
  pages={75545--75559},
  year={2018},
  publisher={IEEE}
}

@article{lian2024tdanet,
  title={TDANet: Target-Directed Attention Network For Object-Goal Visual Navigation With Zero-Shot Ability},
  author={Lian, Shiwei and Zhang, Feitian},
  journal={arXiv preprint arXiv:2404.08353},
  year={2024}
}

@inproceedings{wortsman2019learning,
  title={Learning to learn how to learn: Self-adaptive visual navigation using meta-learning},
  author={Wortsman, Mitchell and Ehsani, Kiana and Rastegari, Mohammad and Farhadi, Ali and Mottaghi, Roozbeh},
  booktitle={Proceedings of the IEEE/CVF conference on computer vision and pattern recognition},
  pages={6750--6759},
  year={2019}
}

@inproceedings{yokoyama2024vlfm,
  title={Vlfm: Vision-language frontier maps for zero-shot semantic navigation},
  author={Yokoyama, Naoki and Ha, Sehoon and Batra, Dhruv and Wang, Jiuguang and Bucher, Bernadette},
  booktitle={2024 IEEE International Conference on Robotics and Automation (ICRA)},
  pages={42--48},
  year={2024},
  organization={IEEE}
}

@article{sethian1996fast,
  title={A fast marching level set method for monotonically advancing fronts.},
  author={Sethian, James A},
  journal={proceedings of the National Academy of Sciences},
  volume={93},
  number={4},
  pages={1591--1595},
  year={1996},
  publisher={National Acad Sciences}
}

@article{schulman2017proximal,
  title={Proximal policy optimization algorithms},
  author={Schulman, John and Wolski, Filip and Dhariwal, Prafulla and Radford, Alec and Klimov, Oleg},
  journal={arXiv preprint arXiv:1707.06347},
  year={2017}
}

@misc{cai2025clcotnavclosedloophierarchicalchainofthought@clcotnav,
      title={CL-CoTNav: Closed-Loop Hierarchical Chain-of-Thought for Zero-Shot Object-Goal Navigation with Vision-Language Models}, 
      author={Yuxin Cai and Xiangkun He and Maonan Wang and Hongliang Guo and Wei-Yun Yau and Chen Lv},
      year={2025},
      eprint={2504.09000},
      archivePrefix={arXiv},
      primaryClass={cs.RO},
      url={https://arxiv.org/abs/2504.09000}, 
}

@article{lin2025navcot@navcot,
  title={Navcot: Boosting llm-based vision-and-language navigation via learning disentangled reasoning},
  author={Lin, Bingqian and Nie, Yunshuang and Wei, Ziming and Chen, Jiaqi and Ma, Shikui and Han, Jianhua and Xu, Hang and Chang, Xiaojun and Liang, Xiaodan},
  journal={IEEE Transactions on Pattern Analysis and Machine Intelligence},
  year={2025},
  publisher={IEEE}
}

@misc{wu2024voronavvoronoibasedzeroshotobject@voronav,
      title={VoroNav: Voronoi-based Zero-shot Object Navigation with Large Language Model}, 
      author={Pengying Wu and Yao Mu and Bingxian Wu and Yi Hou and Ji Ma and Shanghang Zhang and Chang Liu},
      year={2024},
      eprint={2401.02695},
      archivePrefix={arXiv},
      primaryClass={cs.RO},
      url={https://arxiv.org/abs/2401.02695}, 
}

@misc{qiao2025opennavexploringzeroshotvisionandlanguage@opennav,
      title={Open-Nav: Exploring Zero-Shot Vision-and-Language Navigation in Continuous Environment with Open-Source LLMs}, 
      author={Yanyuan Qiao and Wenqi Lyu and Hui Wang and Zixu Wang and Zerui Li and Yuan Zhang and Mingkui Tan and Qi Wu},
      year={2025},
      eprint={2409.18794},
      archivePrefix={arXiv},
      primaryClass={cs.RO},
      url={https://arxiv.org/abs/2409.18794}, 
}

@inproceedings{Liang2020AFBURR,
  author = {Liang, Yongqing and Li, Xin and Jafari, Navid and Chen, Jim},
  booktitle = {NeurIPS},
  title = {Video Object Segmentation with Adaptive Feature Bank and Uncertain-Region Refinement},
  year = {2020}
}

@inproceedings{li2020fastGlobalContext,
  title={Fast Video Object Segmentation using the Global Context Module},
  author={Li, Yu and Shen, Zhuoran and Shan, Ying},
  booktitle={ECCV},
  year={2020}
}

@inproceedings{oh2019videoSTM,
  title={Video object segmentation using space-time memory networks},
  author={Oh, Seoung Wug and Lee, Joon-Young and Xu, Ning and Kim, Seon Joo},
  booktitle={ICCV},
  year={2019}
}

@inproceedings{cheng2021mivos,
  title={Modular Interactive Video Object Segmentation: Interaction-to-Mask, Propagation and Difference-Aware Fusion},
  author={Cheng, Ho Kei and Tai, Yu-Wing and Tang, Chi-Keung},
  booktitle={CVPR},
  year={2021}
}

@inproceedings{seong2020kernelizedMemory,
  title={Kernelized Memory Network for Video Object Segmentation},
  author={Seong, Hongje and Hyun, Junhyuk and Kim, Euntai},
  booktitle = {ECCV},
  year={2020}
}

@inproceedings{lu2020videoGraphMem,  
 title={Video Object Segmentation with Episodic Graph Memory Networks},  
 author={Lu, Xiankai and Wang, Wenguan and Martin, Danelljan and Zhou, Tianfei and Shen, Jianbing and Luc, Van Gool},  
 booktitle={ECCV},  
 year={2020}  
}

@inproceedings{hu2021learning,
  title={Learning Position and Target Consistency for Memory-based Video Object Segmentation},
  author={Hu, Li and Zhang, Peng and Zhang, Bang and Pan, Pan and Xu, Yinghui and Jin, Rong},
  booktitle={CVPR},
  year={2021}
}

@inproceedings{lai2020mast,
  title={MAST: A memory-augmented self-supervised tracker},
  author={Lai, Zihang and Lu, Erika and Xie, Weidi},
  booktitle={CVPR},
  year={2020}
}

@incollection{atkinson1968human,
  title={Human memory: A proposed system and its control processes},
  author={Atkinson, Richard C and Shiffrin, Richard M},
  booktitle={Psychology of learning and motivation},
  volume={2},
  pages={89--195},
  year={1968},
  publisher={Elsevier}
}

@misc{shi2025memoryvlaperceptualcognitivememoryvisionlanguageaction,
      title={MemoryVLA: Perceptual-Cognitive Memory in Vision-Language-Action Models for Robotic Manipulation}, 
      author={Hao Shi and Bin Xie and Yingfei Liu and Lin Sun and Fengrong Liu and Tiancai Wang and Erjin Zhou and Haoqiang Fan and Xiangyu Zhang and Gao Huang},
      year={2025},
      eprint={2508.19236},
      archivePrefix={arXiv},
      primaryClass={cs.RO},
      url={https://arxiv.org/abs/2508.19236}, 
}

@misc{cheng2022xmemlongtermvideoobject,
      title={XMem: Long-Term Video Object Segmentation with an Atkinson-Shiffrin Memory Model}, 
      author={Ho Kei Cheng and Alexander G. Schwing},
      year={2022},
      eprint={2207.07115},
      archivePrefix={arXiv},
      primaryClass={cs.CV},
      url={https://arxiv.org/abs/2207.07115}, 
}

@inproceedings{wang2021swiftnet,
  title={SwiftNet: Real-time Video Object Segmentation},
  author={Wang, Haochen and Jiang, Xiaolong and Ren, Haibing and Hu, Yao and Bai, Song},
  booktitle={CVPR},
  year={2021}
}

@inproceedings{cao2025cognav@cognav,
  title={Cognav: Cognitive process modeling for object goal navigation with llms},
  author={Cao, Yihan and Zhang, Jiazhao and Yu, Zhinan and Liu, Shuzhen and Qin, Zheng and Zou, Qin and Du, Bo and Xu, Kai},
  booktitle={Proceedings of the IEEE/CVF International Conference on Computer Vision},
  pages={9550--9560},
  year={2025}
}
}

\end{document}